# Lessons Learned: The Evolution of an Undergraduate Robotics Course in Computer Science

R. Pito Salas[0000-0002-0788-3614]

Brandeis University, 415 Waltham, Massachusetts, USA
`rpsalas@brandeis.edu`

**Abstract.** Seven years ago (2016), we began integrating Robotics into our Computer Science curriculum. This paper explores the mission, initial goals and objectives, specific choices we made along the way, and why and outcomes. Of course, we were not the first to do so. Our contribution in this paper is to describe a seven-year experience in the hope that others going down this road will benefit, perhaps avoiding some missteps and dead-ends. We offer our answers to many questions that anyone undertaking bootstrapping a new robotics program may have to deal with. At the end of the paper, we discuss a set of lessons learned, including striking the right balance between depth and breadth in syllabus design and material organization, the significance of utilizing physical robots and criteria for selecting a suitable robotics platform, insights into the scope and design of a robotics lab, the necessity of standardizing hardware and software configurations, along with implementation methods, and strategies for preparing students for the steep learning curve.

## 1 Introduction

Introducing robotics into the undergraduate Computer Science curriculum creates distinct challenges. In this paper, we will analyze our experience over seven years, share the plans and challenges we encountered, report on what went well and what did not, put our experience in the context of what others have reported, and extract specific lessons learned during this journey.

### 1.1 Original Motivation

While our Computer Science department offered a robust curriculum covering both theory and practice, we felt there was a gap in preparing our students for life after graduation. In most cases, even when actual programming is part of the syllabus, it would involve individual projects or heavily scaffolded homework assignments that could be completed over a week or two.

In contrast, we understood that whatever our students chose to do with their Computer Science degree, they would quickly find themselves in a context that could be foreign to them: they would likely be working on systems far larger than they encountered in their studies, with a history of previous contributors, and most likely work with a team of members who join and exit. We are not the first to observe this gap; for example, [2018-Gutica] and [2020-Meah].

Based on this we decided to create a new course, specifically around a large, collaborative, multi-semester project, which would better reflect what most of our students would encounter after graduation. Note that at this point, we had not yet chosen robotics as the focus area.

### 1.2 Our Approach: Robotics as part of Computer Science

What kind of course would it be? What kind of project? We chose robotics because the topic not only aligns with current technological trends and student interests, but also provides a practical application of advanced computer science sub-disciplines such as computer vision, machine learning, computational linguistics, and core algorithms.

Our vision was to develop a program that went beyond the theory, also enabling students to produce tangible outcomes, such as sophisticated projects, which they could showcase in various formats, including online portfolios and resumes.

### 1.3 Putting our experience in context

Creating new robotics courses and programs has been an area of innovation for the last 20 years, and is reflected in the literature ([1998-Thangiah], [2011-Aliane], [2018-Raman], and many others). Even when we started, there had been a growing number of similar courses and programs, and since then, the numbers have only increased.

A survey conducted by Shibata et al. [2021-Shibata] found 14 Universities with a Robotics Department in Japan alone. While this paper focuses on graduate programs, it presents a view of the evolution of robotics courses and programs. It seems that many robotics departments in this study have grown out of hardware, electrical, mechanical, and engineering disciplines, and so have a far heavier focus on the hardware compared to our focus, which was to be the software of robots.

### 1.4 Program Goals and Structure

To implement this vision, the plan was to create a multi-semester program around our pilot course, covering autonomous robotics from a theoretical and applied perspective.

Therefore, hands-on learning should be prioritized, with programming assignments designed to instill basic principles, theories, and techniques of robotics, alongside practical experience with physical robots and software engineering.

To ensure the course's future value, it would be important to ensure that it was engaging, rewarding, and beneficial for their forthcoming careers as software engineers or computer scientists, irrespective of whether they pursue robotics further. This approach aimed to mirror the challenges and environments that students would likely encounter after graduation.

Our success in meeting these goals is evaluated in our concluding remarks.

## 1.5 Related Work

Over the past 20 years, there have been numerous reports of fledgling robotics courses and programs, such as [1998-Thangiah], [2018-Pieters]. Of course, even ten years is a lifetime when it comes to computing and robotics. Therefore, it is worth considering the past and reevaluating the present and future in the context of change. We have organized our review by topic.

*Curriculum Design*: There is a large amount of literature on the basic question of designing the curriculum of a robotics course. We took a lot of inspiration from Touretzky's Seven Big Ideas paper [2012-Touretzky], that presents "seven *big ideas* in robotics that can fit together in a one semester course for junior or senior level computer science majors." Raman et al [2018-Raman] describes the design of a graduate level course for engineering students where they cite very similar choices and decisions. Interestingly, they also cite Touretzky 'swriting that "the framework for teaching was closely designed in the format suggested by Touretzky".

*Breath vs. Depth and overall scope* - How do we cover the very broad topic of "Robotics," especially the first course? Therefore, there is a real risk of trying to pack too much into a single course. Nikolaus Correll identifies four key challenges in "A One-Year Introductory Robotics Curriculum for Computer Science Upperclassmen" [2013-Correll]: scalability, student assessment, depth vs. breath. They say: "This viewpoint positions "robotics" not as a new discipline, but as an extension of computer science and its sub-disciplines including artificial intelligence, natural language processing, computer vision, software engineering, security, and human-computer interaction, among others." Michaelis and Weintrop link Computer Science Pedagogy to the Interest Development Literature [2022-Michaelis]. Several themes emerge in this work: the importance of scaffolding that is appropriate to the students' current mastery and the importance of connecting what is being taught with the students' real-world experiences and goals.

*Importance of hands-on learning*–It is an almost accepted dogma that hands-on learning and active learning are highly beneficial in almost any kind of learning. For example, in the review article "Does Active Learning Work? A Review of the Research" [2004-Prince], Prince concludes: "It is found that there is broad but uneven support for the core elements of active, collaborative, cooperative and problem-based learning."

*Use of ROS in the educational Context* - An important question when designing a robotics course or program is what framework should be adopted. Pieters, Ghabcheloo, and Lanz conclude that "This approach leads to much trial-and-error and a steep learning curve that, we believe, is highly valuable. We reflect on an early evaluation of our approach and conclude that the advantage of ROS as tool to educate robotics is due to its holistic nature."[2018-Pieters]. Using ROS also prepares our students for future jobs in robotics space. As Raman et al [2018-Raman], cited earlier, states: "Now such knowledge and understanding of ROS is already a key requirement for the majority of open positions in the Autonomous Vehicle R&D industry."

## 2 Course Structure

The key goal was to build a course around a challenging, long-term objective. Our initial choice, "The Campus Rover," which would be a delivery robot which would deliver packages between offices on campus, including moving both indoors and outdoors, locate offices automatically, even navigate to different floors.

This was the theme of the course and the stated objectives of the associated laboratories. We naively believed this was a "solved problem" which we just had to replicate.

We first describe this approach as "Multi-semester-Multi-Cohort" [2019-Salas], "The course structure is framed around a long-term vision across multiple semesters, with different groups of students each semester. We know that the vision is ambitious and that it will take more than one semester to achieve."

### 2.1 Progress

Our initial course, "Autonomous Robotics," is a standard one semester course, running 13 weeks, with two 80 minute "lectures" and one 2-hour Lab each week. There were five individual programming assignments and one major term project for teams of two to four students.

We offered the course for the first time in 2016 and it has run every year since then. The chart below shows that some years the course was offered both in the Spring and Fall. As the course has become better known, enrollment has increased with a significant proportion of Grad students.

In course evaluations, the course never received a rating lower than 4.3/5, which is interesting but difficult to put into context, as departmental and university-wide statistics are not available.

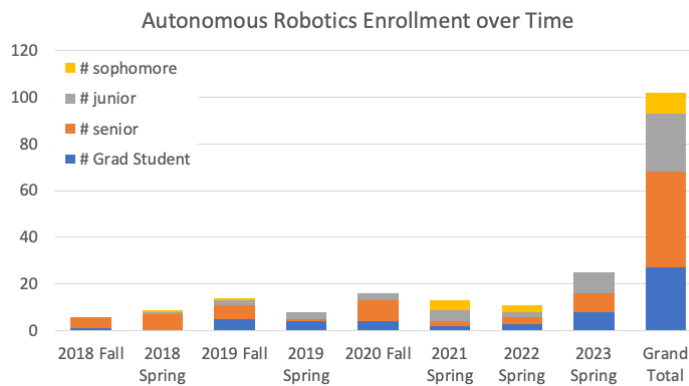

### 2.2 Syllabus

The syllabus evolved over multiple iterations of the course, just as the objectives evolved. The objective was to review the most important ideas in autonomous robotics.

For this we took a lot of inspiration from Touretzky's Seven Big Ideas paper [2012-Touretzky], in which Touretzky states: "we present seven big ideas in robotics that can fit together in a one semester undergraduate course. Each is introduced with an essential question, such as "How do robots see the world?"

It is possible to quibble with the particular choice of "Big Ideas," but the concept is sound and powerful. Our syllabus has evolved over time. It has been, and continues to be, an open source and publicly available [2024-Salas].

This section provides a brief overview of the syllabus. As shown, we started by teaching students ROS and how to work with physical robots. The next series of lectures and labs covered the "Seven Big Ideas. Finally, the course touches on some advanced concepts (Kalman Filters, Behavior Trees) that are not strictly core. From our experience, they are used in many students' projects.

| Welcome | ROS Basics (2) | How are robots built? | How do Robots Sense their environment? (2) | How do mobile Robots move? | ROS Basics, Continued |
|---|---|---|---|---|---|
| How does a robot know its orientation? | Coordinates and transforms (2) | How does a robot know where it is? (2) | How does a robot decide where to go? (3) | How does a robot decide what to do? (3) | How does a robot interpret what it sees? (2) |
| How does a robot deal with noisy sensors? | How do robots manipulate things? | Project work. | Project work. | Project work. | Project work. |

**Figure 1.**

Autonomous Robotic Course Syllabus. Read weeks, left-to-right, top-down. The numbers in (parentheses) are topics requiring multiple weeks.

### 2.3 Term Projects

This aspect of our structure has significantly evolved in recent years. As previously mentioned, the original plan for term projects was to build a long-term project, the "Campus Rover."

This was consistent with our original vision, but there were challenges: Not all students were excited about working on the project. They were more interested in their own creative ideas. In addition, we realized that our robot (Robotis Turtlebot3) was not suitable because it was not outdoor-capable. We substantially underestimated the complexity of the goals set for this project. After the first three cohorts, we pivoted to allow open-ended project ideas from the students, which resulted in more buy-in.

## 2.4 Continuity

As mentioned, it was important to give students the experience of working on a project that others had worked on, and that others would work on again in the future. This notion was kept front and center throughout: "your code is not throw-away; your learning is not throw-away. Indeed, we are building a kind of "institutional memory" that will last years.

**"Dear Future Student" Letter:** Many projects lead to the creation of tools, software, techniques, or data that could be used by another project. In such cases, students are asked to write a "Dear Future Student" letter [Salas-2024-2], who is asked to reflect on the next student or user of what they contributed and what they would want to have if they were to pick up the project. Framing it as a "letter" again reinforces the sense of continuity and responsibility. Often the letters are personal and encouraging.

**Lab Notebook:** Our program incorporates a detailed and open-source laboratory notebook [2024-Salas-4]. This resource allows instructors to disseminate essential procedures, rules and infrastructure-related information. More significantly, it functions as an extensive repository of accumulated knowledge, structured akin to comprehensive FAQ.

Each student was required to contribute two original FAQ entries at the conclusion of the term, reflecting on the challenges they encountered and their resolutions. These contributions vary from algorithmic methodologies and hardware functionalities to newly discovered software tools, collectively enriching this notebook as both an academic reference and chronicle of the program's evolution.

Although this system has proven invaluable, it is not without its limitations. The quality and depth of student contributions vary, and occasionally the solutions presented may not represent the most effective approach. Additionally, with the rapid advancement of technology, certain entries may become outdated. Nonetheless, the Laboratory Notebook remains a pivotal tool in fostering a sense of community, continuity, and collective responsibility towards the program's overarching mission.

**Code Repository:** There is a single major code repository for all assets built by instructors and students, including final projects, special projects, and software tools. [2024-Salas-3] This serves the same dual purpose as the Letter and the Lab Notebook: it creates a continuing institutional memory of learning and work done and simultaneously engenders a sense of continuity and responsibility to the next person or team.

## 2.5 Robotics Learning Lab

Our experience is that having a dedicated Robotics Lab, even if it is tiny, is almost mandatory. The lab serves as a place for students to work on their projects and test them, and a place to store robots and other tools, parts, and equipment when the class is not actively running. However, the lab's role in creating a sense of belonging and responsibility for the process is important.

**Working without a lab.** Sometimes, a dedicated laboratory is not possible. In this case, there are two less desirable options. One could teach robotics purely in simulation. Indeed, as discussed elsewhere, working with simulators is an essential skill to learn, but working exclusively through simulation is not recommended. Also possible is a temporary lab. This involves setting up and tearing down a lab area as needed; while this can be made to work, it is clearly not ideal. Over time, a Robotics Lab will accumulate many tools, equipment, robots, batteries, etc..

## 2.6   Selecting a Robot for Teaching Robotics

Choosing the right robot platform is crucial, as it influences the curriculum, assignments, student experience, and budget. Our experience with various robots, both commercial and custom-built, informed us of our priorities.

**Cost-effectiveness:** We aim for each student or team to have their own robot, keeping costs under $1,000 each, ideally around $500, to support a program size of 30-50 students annually.

**Robot Operating System (ROS):** We standardized on ROS for its broader educational and research community support (see next section.)

Open Architecture: Essential for integration flexibility. While most commercially available robots use common single-board computers, such as Arduino, Raspberry Pi, and their variants, they are often paired with proprietary firmware proprietary components.

Although this simplifies construction and improves reliability, it is less customizable. To adapt to dynamic classroom needs, robots must support additional sensors and actuators that require sufficient space and power sources.

**Outdoor Capability:** Robots need large wheels for diverse outdoor terrains, aligning with the curriculum's focus on outdoor navigation.

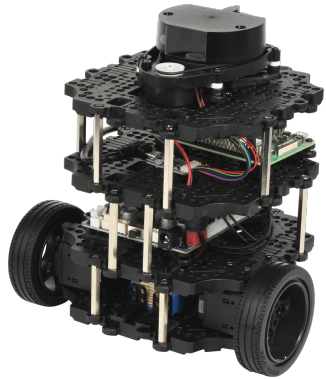

## 2.7   The Robots

As of 2024, the TurtleBot3 from Robotis is available for around $700 [Web-Robotis]. It is a very effective ROS robot. It is well-engineered, reliable, and has excellent documentation. We have had a great deal of success with this.

However, with the small wheels, it would not work outdoors.

Our second major robot is the homegrown "Platform" line of robots.

They are based on software from the open source Linobot3 project [Web-Linorobot]. Our design features large wheels and powerful motors for outdoor use. And it is extensible as can be seen here with an optional arm for gripping objects.

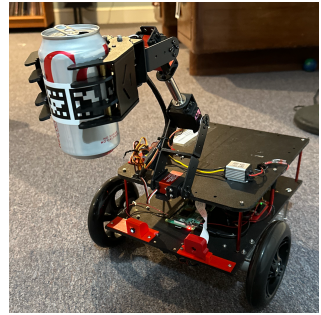

## 2.8 Why ROS (Robotics Operating System)

Choosing a robot and an operating system goes hand-in-hand, which is why we refer to a robot platform. This question has been addressed by various authors in the literature, as mentioned earlier and in [2014-Michieletto]. While ROS is certainly not the only "professional' robotics platform, it is well known and respected. Roboticists in both academia and industry are familiar with ROS, supporting the goals of future value to students.

Once we decided to use the Robotics Operating System (ROS), we confronted several challenges:

**Inherent Complexity.** ROS is a distributed operating system with a complex programming model. It has a very steep learning curve. Robotic algorithms are inherently real-time and asynchronous, sensors are noisy, and hardware is quirky. There is no easy way to overcome this problem, which is an essential part of learning.

**Student Development Environment.** All our students had laptops, but on average, they did not meet the requirements for developing robotics software (due to the amount of memory, free storage, battery life, and speed). To overcome this we created a cloud based ROS environment with Kubernetes on in house servers

## 2.9 Safety, Security and Breakage

To encourage student commitment and satisfaction, one would like to have a dedicated lab that is always open and accessible. (24 × 7). However, there are some practical considerations.

The Lab holds expensive robots and other assets that need to be protected from loss or theft. Therefore, it is critical to control access to the laboratory and to have a good inventory of what is in it. The Robots and ancillary equipment are delicate. Physically, they can be dropped, stepped on, or otherwise mishandled.

**Budget and Cost:** As should be clear from this report, creating a Robotics course and program should not be undertaken lightly. We were in a bootstrap situation and were able to achieve a lot with minimal budget. Our own experience is that the key cost drivers are the price of the robots and the number of students. The following are the guidelines based on our experience:

|  | Basic | Comfortable |
|---|---|---|
| Students per Robot | 5 | 2 |
| Cost per Robot | $500 | $2,000 |
| Students per teaching assistant | 20 | 10 |
| Tools, Parts, Equipment per robot | $200 | $500 |

**Table 1.** Sizing Guidelines

## 3   Lessons Learned and Conclusions

This paper reviewed the genesis of the robotics course and program, explaining the original motivation and goals, the unique approach, choices we made, and the specifics of our course structure and design. The program has evolved, changed, and continued to change.

1. **Lesson: Prepare for a steep learning curve.** Robotics is challenging to teach and challenging to learn. Working with physical computers, sensors and motors teaches you quickly that real-time programming is unpredictable and difficult to make repeatable. We need to prepare students for this, and yet that is from where a large part of the learning comes.
2. **Lesson: Be thoughtful about choosing what to cover.** Robotics is an arbitrarily broad topic. Our finding is that autonomous robotics should be the core, and covering the "Seven Big Ideas" is an excellent way to organize the content.
3. **Lesson: Working with physical robots is essential.** Combining theory and working with physical robots is highly beneficial and yet adds complexity. We have found that while one can learn some of the theory with simulation, there is a dimension of "real robotics" come through only when working with real robots.
4. **Lesson: Standardize on one robot platform.** Physical robots are flaky. Standardize the model and software installation to the maximum extent possible. Our selections were small mobile robots with a camera and a lidar sensor.
5. **Lesson: Do not try to use students' own laptops:** Our experience is that running the software needed to program robots is very resource intensive, computer specific, and sensitive to configuration, and will be a serious time sink. Instead, we recommend providing a standardized environment.
6. **Lesson: Provide a standard environment.** A Linux-based cluster that is accessible over the Web is an excellent solution. They can be home-grown or provided by a commercial vendor. Our cluster runs on four modest servers and simultaneously performs well for approximately 20 users.
7. **Lesson: Minimal Infrastructure:** Minimal infrastructure is a cloud-based simulation environment with one robot for every four students. It is important to have enough robots to acquire fewer higher-end robots. It is better to acquire more, lower-end robots, than fewer higher-end robots. The student-to-robot ratio should be as low as possible.

8. **Lesson: Things are always harder than they appear.** Students always attempt projects that are far more challenging than the time allows. It is crucial to review their plans ahead of time and manage their expectations.

## 3.1 Conclusions

Finally, we return to our original goals and review the results.

**Multi-semester.** The course would be specifically designed around a large, collaborative, multi-semester project. *Assessment*: This was originally one of our motivations for creating a course that would mimic a real-world project that spanned more than a single semester. We established a course where each cohort was built on the previous one, and the results were shared and improved upon. However, this approach was only partially successful. While the idea appeals to students in principle, we found that allowing students creative leeway in selecting a term project was also a great motivator. We believe in the model and continue to work on it.

**Pilot Course.** We created a one-semester "pilot" course in Robotics for our Computer Science students. *Assessment*: While that part was a success, we did not succeed in adding more courses or faculty members. This is explained in part by budgetary constraints, but also a matter of focus of the department.

**Theory and Practice.** The course would teach the fundamentals of Robotics with a clear balance of theory and practice. *Assessment*: Designing a one-semester course covering "robotics" is nearly impossible. The course has evolved to be one of applied robotics. We conclude that one semester is not sufficient to achieve a good balance.

**Hands-on.** The course would include programming assignments from the start, as a major vehicle for learning. *Assessment*: The assignments have continued to evolve and have fulfilled this goal. Feedback from the students was uniformly positive. One challenge is the sequencing of the coverage of concepts with the timing of programming assignments.

**Future Value.** It would be rewarding for students and valuable for their future success. *Assessment*: Only approximately 20% of the students chose to continue in robotics. However, the feedback shows that the great majority of students feel that having taken the course contributed positively to their future career plans.